# Bayes Networks for Sonar Sensor Fusion

**Ami Berler** and **Solomon Eyal Shimony**
{ami,shimony}@cs.bgu.ac.il
Dept. of Math. and Computer Science
Ben Gurion University of the Negev
P.O. Box 653, 84105 Beer-Sheva, Israel

## Abstract

Wide-angle sonar mapping of the environment by mobile robot is nontrivial due to several sources of uncertainty: dropouts due to "specular" reflections, obstacle location uncertainty due to the wide beam, and distance measurement error. Earlier papers address the latter problems, but dropouts remain a problem in many environments. We present an approach that lifts the overoptimistic independence assumption used in earlier work, and use Bayes nets to represent the dependencies between objects of the model. Objects of the model consist of readings, and of regions in which "quasi location invariance" of the (possible) obstacles exists, with respect to the readings. Simulation supports the method's feasibility. The model is readily extensible to allow for prior distributions, as well as other types of sensing operations.

## 1 INTRODUCTION

Sonar mapping of the environment by mobile robot is nontrivial due to several sources of uncertainty. The first impediment to mapping is that most sonars used in mobile robots are wide-beam, in order to allow for sufficient coverage for obstacle avoidance. The wide angle causes a large uncertainty of location of the detected obstacle(s), within the arc at the detected distance from the sonar (such as region $A_1$ in figure 1). Second, readings are inexact due to noise and pulse-width problems (thickness of region $A_1$). These problems have been largely addressed in work by Elfes using a Bayesian model and an occupancy grid [3].

Third, surfaces of objects slanted far from the perpendicular to the direction of the beam may reflect almost no acoustic energy in the direction of the sensor (which is also the emitter - the sonar(s) mounted on the robot). We call such a false negative a dropout, although sometimes, rather than generating no reading at all, the beam, after one or more secondary reflec-tions, is detected by the sonar (a spurious reading) and results in a phantom obstacle. This problem becomes acute if there are many flat surfaces, which are acoustically "specular", i.e. reflect almost all acoustic energy in a narrow cone around the angle of incidence, resulting in frequent dropouts, as has been encountered by various researchers and implementers [12]. The "specularity" of a surface (determined by material, surface roughness, etc.) is usually represented as a "critical angle" from the perpendicular, beyond which the reading is likely to be bogus (in fact the probability that the reading is bogus may depend on other factors, such as signal strength, area, etc., but we would like to abstract away from these issues in this paper).

Most schemes that perform sonar mapping consider information from numerous sonar readings in order to overcome the errors. In a mapping architecture developed by Elfes [3], (2D) space is partitioned into a grid, where each grid point can either be occupied or free. Each grid point is seen as a random variable, denoting probability of occupancy. Information is accumulated, for each grid point, from multiple sonar readings and robot locations (the latter appear as circles in figure 5), resulting in an occupancy probability map as in figure 7 (where black stands for high occupancy probability). An error model of the sonar reading, that handles mostly range error due to noise and pulse width, is used. In general, Elfes makes the following independence assumption: all grid points are independent random variables. Updating the occupancy probability $P(O)$ of each grid point is done according to the sensor reading model $p(r|O), p(r|\neg O)$, *independently from updates to other grid points*, as follows[1]:

$$P(O_{new}) = \frac{p(r|O)P(O_{old})}{p(r|O)P(O_{old}) + p(r|\neg O)(1 - P(O_{old}))} \quad (1)$$

Another system [12] also uses a grid-based scheme, but with Dempster-Shafer probabilities [2]. Each grid point (cell) has the values $h_o$ and $h_f$, denoting belief that the cell is occupied, and belief that the cell is free, respectively. Update of the model is, again, indepen-

---

[1] We denote the probability density function by $p$, and discrete probability distribution by $P$ throughout.

dent in each cell, as follows[2]:

$$h_f^{new} = \frac{1}{\delta}(h_f^{old}h_f^{obs} + h_f^{old}h_u^{obs} + h_u^{old}h_f^{obs})$$

$$h_o^{new} = \frac{1}{\delta}(h_o^{old}h_o^{obs} + h_o^{old}h_u^{obs} + h_u^{old}h_o^{obs})$$

with $h_u = 1 - h_o - h_f$, and the superscripts *old, new, obs* denoting old belief values, new belief values, and observed values (by the current sensing operation), respectively. $\delta$ is a "normalizer", used to account for the probability mass of contradictory evidence, as follows:

$$\delta = 1 - (h_o^{old}h_f^{obs} + h_f^{old}h_o^{obs})$$

Dropouts (false negatives) are handled by direct filtering on successive readings during travel, and the problem of uncertainty due to wide beam is alleviated by the facts that: a) The sonar beam width of only about $10^0$, compared to the roughly $25^0$ beam width of the standard Polaroid sonars. b) Triangulation between readings from different robot locations (also facilitated by the narrower beam). Remaining errors are handled by the updating mechanism above. One disadvantage of the scheme is obviously the larger number of readings required due to the narrower beam. The paper [12] does not give a clear semantics to the prior belief values for the cells. It is also unclear whether the method is applicable to a more cluttered environment than the rather simple one in the experiment, or to objects with a more specular reflectance pattern.

In essence, both schemes assume that grid points are independent variables, both a-priori *and* given the sensing operations. There are several problems in assuming independence of grid points and of not keeping track of which readings originated the information. Consider, for example, the spatial configuration of readings in figure 1. In the figure, the sectors are regions where, had there been an obstacle, the readings would have been other than observed (assuming no false negatives). The thickened arcs are regions where some object is likely to be (as a cause for the reading value). Using grid-point independence, reading $R_2$ has no effect on the occupancy probability in region $A_1$, whereas the fact that a possible obstacle in $A_1S_2$ is contra-indicated by $R_2$, and a possible obstacle in $A_1A_2$ is confirmed by $R_2$, should change the probability that region $A_1$ is occupied.

Increased probability of occupancy in region $A_1A_2$, and decreased probability of occupancy in region $S_1S_2$ occur both in Elfes' model and our own. Extending the example, consider a reading $R_3$ that might be caused by an object in region $A_3$, which partially overlaps $S_1$, as in figure 2. The probability that $R_1$ is a dropouts increases, and this should decrease the probability of occupancy for region $A_1$. This kind of inference cannot be handled by a scheme that assumes independence of grid points, and does independent updates.

---

[2]In the cited paper, $h$ values are actually in the range 0 to 100, but that is immaterial.

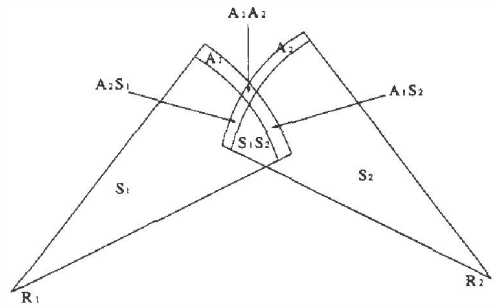

Figure 1: Spatial Configuration of 2 Sonar Readings

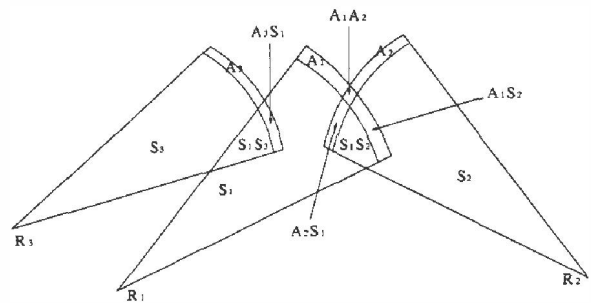

Figure 2: Spatial Configuration of 3 Sonar Readings

We do *not* believe that the difficulty, exhibited by earlier Bayesian schemes (e.g. [3]) in handling dropouts, results from the fact that the model is Bayesian. Rather, as stated earlier, we hold that this is due to over-optimistic independence assumptions. Additionally, the fact that sensor models are usually given in terms of single probability distributions, favors use of the Bayesian scheme. Our model is based, in essence, on Elfes' model, but without assuming that grid points are independent, thus overcoming the above problems. In our scheme, space is dynamically partitioned into regions (depicted in figures 1, 2 for the above examples), dependent on the readings. (In our scheme, we do not necessarily need to quantize space into a grid, as in prior work.) Hence, our model partitions the plane into regions which are "quasi-invariant" with respect to the set of readings. That is, if an obstacle is within a region, moving it elsewhere within the region (or even adding and removing obstacles from the region, as long as it is still partially occupied) makes "almost" no difference in the readings. Each such region is a potential cause for a set of readings, and may indicate that some other set of readings are dropouts, as discussed in the above examples.

The remainder of this paper is organized as follows. Section 2 explains our assumptions, presents the model, and its realization as a Bayes network [9]. Section 3 discusses our implementation of the model. Section 4 presents initial simulation results. We conclude with notes on how to extend the model to handle more complicated prior models and other types of sensors.

## 2    THE SENSING MODEL

We first present and justify our single sonar reading
sensor model assumptions, and then discuss the issue
of independence assumptions we have remaining in the
multi-reading sensing model.

### 2.1    Assumptions and Definitions

Our single reading model assumes the following:

1. The probability that an obstacle is detectable by
   a reading depends only on whether it is (at least
   in part) in a sector in front of the sonar, defined
   by the beam-width. Probability of detecting an
   object outside the sector is 0.

2. Given a true reading $r$, the probability of occu-
   pancy of each unit area in the thickened (thick-
   ness $\epsilon$, the reading inaccuracy) arc about distance
   $r$ from the sensor is uniform. The probability
   that there is an object in the beam at a distance
   $d < r - \epsilon$ is 0.

3. Spurious readings are uniformly random.

4. The probability of a reading given an object in
   the thickened (detection) arc is independent of its
   location within the arc[3].

All the above assumptions are approximations to a
more realistic sensor model, made so that we can par-
tition space into equi-distributed regions. Assumption
2, uniform distribution (in contrast with the realis-
tic model, of a normal distribution along the distance
axis), is reasonable if the error of a true reading is small
compared to other errors (such as arc length). In many
sonar systems, where distance error is on the order of
1 inch, the above condition holds. Assumption 3 is
very difficult to overcome without further spatial data,
which is unavailable in an unmapped environment, but
we believe that the results are not very sensitive to its
violation.

In order to define a probability distribution over a
set of readings, several independence assumptions are
used, as follows:

1. Readings are independent measurements.

2. Occupancy of regions are (a-priori) independent.

While the first assumption is commonly used, the sec-
ond is more problematic. Prior work (e.g. [3]) makes
this assumption w.r.t. individual occupancy grid cells.
Doing so w.r.t. regions should present no difficulty. In

fact, it should be advantageous, since one can con-
sider the size of a region, as well as its shape, when
determining prior probabilities of occupancy. How-
ever, since regions are arbitrary, and modified dynam-
ically, this is equivalent to assuming independence of
grid points. Later on, we adjust this assumption. The
main difference with prior work is that region occupan-
cies are *not* independent given the readings, contrary
to Elfes' model.

The resulting probabilistic model for multiple readings
is as follows. Conditional probabilities for each read-
ing are dependent only on the occupancy of regions
within their sector or thickened arc. Let reading $r$ be
a random variable with range $[R_{min}, R_{max}]$, $A_r$ be its
arc region, and $S_r$ be its sector. Let free($X$) be a pred-
icate that denotes that region $X$ is completely free (i.e.
has no obstacles)[4]. Then:

$$\begin{array}{rcl} p(r|\neg\text{free}(S_r)) & = & p_s(r) \\ p(r|\text{free}(S_r) \wedge \text{free}(A_r)) & = & p_0(r) \\ p(r|\text{free}(S_r) \wedge \neg\text{free}(A_r)) & = & p_r(r) \end{array} \qquad (2)$$

where the three cases stand for a dropout in region
$S_r$, a spurious reading due to noise where no objects
exist (such as crosstalk, sensor error, etc.), and a true
reading, respectively. The probability of a reading is
independent of occupancy of any regions outside $S_r$
and $A_r$. We now define the conditional distributions
$p_s(r), p_0(r), p_r(r)$ in the regions of $R_{min} \le r \le R_{max}$.
Outside this range, the distributions are defined arbi-
trarily, to make them normalized.

Under our assumptions, the conditional distribution
$p_r(r)$ is piecewise uniform, with a high value $p_r^h$ at
roughly the distance of the region $A_r$ from the source,
and low value $p_r^l$ elsewhere inside the sonar range. The
exact ratio depends on the prior probability $P_r$ that
the primary reflection from objects in $A_r$ is detected:
$P_r = 2\epsilon p_r^h$. The distribution $p_s(r)$, for dropouts in
region $S_r$, is uniform, under the assumptions. As we
have no good model for other sources of errors, and
have not encountered them in our system, we use a
piecewise uniform $p_0(r)$, with a low value for all pos-
sible readings (i.e. within the sensor range). In fact,
since the reading in our system is discrete, we use a
discrete probability distribution instead of densities.

For each region $X$, the probability of free($X$) is depen-
dent on area, shape, and location (if there is any prior
information). The set of regions used in the model is
the Cartesian product of the distinct regions of all $n$
readings. That is, let $\{A_{r_i}, S_{r_i}, E_{r_i}\}$ be the arc, sector,
and the remaining space for reading $r_i$, respectively.
Then our set of regions, for $n$ readings, is:

$$\mathcal{A} = \prod_{i=1}^{n} \{A_{r_i}, S_{r_i}, E_{r_i}\}$$

Note that although the size of $\mathcal{A}$ seems to be expo-
nential in $n$, there are two mitigating factors: a) In practice, most of the regions are spatially empty regions, and are thus ignored. If all thickened arcs and sectors were convex, the actual maximum number of non-empty regions would be low-order polynomial in the number of readings. The arcs are not convex, but are nearly so, and we would thus expect the number of regions to be comparatively small. b) In the actual implementation we use a discrete representation, and the number of distinct regions is at most equal to the number of grid points. Henceforth, when we refer to $\mathcal{A}$, we exclude the spatially empty regions, and in the implementation actually mean the non-empty regions in the discrete representation of the sets in $\mathcal{A}$.

---

[3]Strictly speaking, taken together with assumption 2,
this entails that readings $r$ and distances of objects from
the sonar $d$ are equi-distributed, which involves compli-
cated spatial assumptions. However, if $\epsilon$ is reasonably
small this effect is negligible, and we thus ignore this con-
straint henceforth.

[4]We denote the random variables standing for the oc-
cupancy status of a region by its name $X$, and the state
that it is completely free by the predicate free($X$).

Using discrete distributions for the readings, the complete distribution is as follows:

$$P(r_1, ... r_n, s(\mathcal{A})) = \prod_{A \in \mathcal{A}} P(s(A)) \prod_{1 \le i \le n} P(r_i \mid s(\mathcal{A})) \quad (3)$$

where $s()$ stands for "state of region(s)" (either free or occupied). In fact, since $r_i$ is independent of regions in $E_{r_i}$, we can condition $r_i$ only on the state of:

$$\mathcal{A}(r_i) = \{A | A \in \mathcal{A} \wedge A \subseteq (A_{r_i} \cup S_{r_i})\}$$

where $P(r_i \mid \mathcal{A}(r_i))$ is defined in equation 2.

The prior probabilities that each region $A$ is free is determined by its area, and prior probability of occupancy of a point in the room. We use the piecewise linear function:

$$P(\neg\text{free}(A)) = \max(p_{min} + k \mid A \mid, p_{max}) \quad (4)$$

$p_{min}$ is the fraction of the space which is occupied, and $p_{max}$ is some probability close to 1. In principle, one could determine the values $k$, $p_{min}$ and $p_{max}$ by sampling and line-fitting. Admittedly, this scheme is somewhat ad-hoc, and a better model could be constructed, but we believe that this scheme is sufficient at this point. Note that, strictly speaking, this model is *inconsistent* with the assumption of independent region occupancies, from a *dynamic* point of view.

If we had a *static* set of readings, and just assumed that regions in $\mathcal{A}$ are independent, the result would be consistent. However, if we now add another reading $r_{n+1}$, thereby splitting some region $A$ into $A_1$ and $A_2$, then assuming that $A_1$ and $A_2$ are still independent is inconsistent with the inverse probability assignment function, since: $P(A) \ne P(A_1)P(A_2)$.

**Theorem 1** *The only region-area based probability assignment consistent with the independence assumption is the* exponential *probability function, i.e.* $P(free(A)) = e^{-c|A|}$

Proof: Clearly, the exponential probability assignment function is consistent with the assumption, since for every bisection of region $A$ into $A_1$, $A_2$, we have:

$$P(\text{free}(A)) = e^{-c|A|} = e^{-c|A_1 \cup A_2|} = e^{-c(|A_1| + |A_2|)} =$$
$$= e^{-c|A_1|}e^{-c|A_2|} = P(\text{free}(A_1))P(\text{free}(A_2))$$

and this holds by an inductive version of this argument for every partition of $A$. Uniqueness of the solution follows immediately from the uniqueness theorem for certain families of differential equations, of which the equation $f(x) = cf'(x)$ is a member. □

However, using an exponential probability function is counterintuitive: using any non-0 value of $c$, we get that the probability for any small region being free approaches 1. This effect occurs because of the unrealistic assumption of region independence, and ideally, one should add some prior (possibly Markov) dependency model. For simplicity at this point, we prefer to slightly (dynamically) violate the independence assumption, by using a non-exponential probability assignment, while still assuming that the assumption holds on each *static* model.

## 2.2 Realization as a Bayes Network

The probabilistic reasoning issue we need to tackle is to find the distribution of region occupancies given the sonar readings. We actually solve a somewhat simpler problem - finding the (marginal) probability that each region is occupied given the readings. Representing the distribution as a Bayes network allows us to use existing tools to solve the problem.

Equation 3 is a conditional distribution where reading probability is dependent only on a set of region variables, and region variables are independent. It thus exhibits a 2-level dependency structure, and can thus be implemented as a 2-level Bayes network, one root node for each region in $\mathcal{A}$, and one sink node for each reading. [5] However, since the in-degree is essentially unbounded, and a reading node is *not* a type of node for which specialized implementations exist in the literature (as is the case for, e.g. noisy-or), we prefer to implement the model as a 3-level network, by adding, for each reading-node $r_i$, two intermediate (binary valued) parents: "pro" and "con", which are or-nodes, defined as follows:

$$P(\text{pro}_{r_i} = true) = \begin{cases} 1 & \exists A \subseteq A_{r_i}, \ \neg\text{free}(A) \\ 0 & \text{otherwise} \end{cases}$$

$$P(\text{con}_{r_i} = true) = \begin{cases} 1 & \exists A \subseteq S_{r_i}, \ \neg\text{free}(A) \\ 0 & \text{otherwise} \end{cases}$$

Additionally, since in the implementation, a reading node is always set to the observed reading, it is sufficient that $r_i$ be a binary-valued node, with one value, "obs", standing for the observed value, and another standing for all other possible readings.

$$P(r_i = \text{obs}|\text{pro}_{r_i}, \text{con}_{r_i}) = \begin{cases} P_{s_{r_i}} & \text{con}_{r_i}, \\ P_{0_{r_i}} & \neg\text{con}_{r_i} \wedge \neg\text{pro}_{r_i} \\ P_{r_{r_i}} & \neg\text{con}_{r_i} \wedge \text{pro}_{r_i} \end{cases}$$

---

[5] Note that readings are assumed independent given the environment, but are *dependent* in an unknown environment. Likewise, regions are (approximately) independent a-priori, but must become dependent given the readings. Hence, the Bayes network must be constructed in the causal direction: regions preceding readings in the DAG.

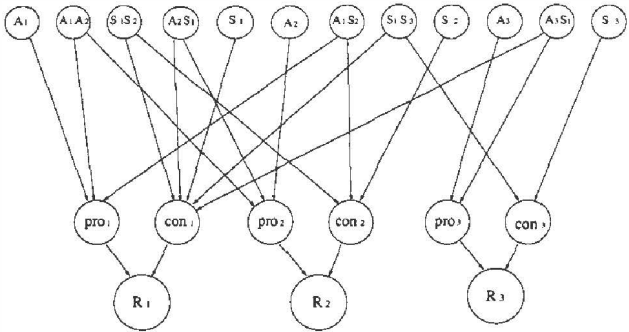

Figure 3: Bayes Net for 3 Sonar Readings

We use a constant low $P_{0_{r_i}} = P_0$ (that is, assume few spurious readings that are *not* dropouts), and $P_{r_{r_i}}$ a constant $P_r$, a reasonably high probability for initial experiments. In fact, this number should be determined by the spatial configuration, but we do not as yet have a good model for determining it. $P_{s_{r_i}} = P_s$ is determined by the prior probability that an object in the sector will cause no detectable primary reflection (i.e. that we will get a dropout)[6].

Each region $A$ is represented by a binary random variable, denoting free($A$), with prior probability as discussed above. The resulting network can now be implemented with existing Bayes network software that handles or-nodes as special cases (see implementation section on using systems with explicit distribution representation). Such networks are similar to BNO2 networks [1], except for the negative-cause con nodes. For example, the Bayes network for the configuration of Figure 2 is depicted in figure 3.

## 3  IMPLEMENTATION

The experimental system has two main components: the readings analyzer, and the probabilistic reasoner (Figure 4). The former communicates with the Nomadic Software Development package (simulator) for the Nomad 200 mobile robot (which can also communicate in turn with the physical Nomad 200 robot).

The readings analyzer is a C++ program consisting of four modules: The robot activation module, the analysis module, the output module and the graphics module. The analyzer uses a data structure with two partitions: dynamic structure for the sonar readings and regions, and dynamic grid/pixels relations with the above structure. Each sonar reading produces two or more new regions. While it is possible to use a purely geometrical representation in our model, the grid was found to be easier and possibly a more efficient way to implement the system in practice.

---

[6]We do *not* divide that probability by the number of possible readings, since the value "observed" plays the role of all other incorrect readings, where there is indeed an object in the sector, which was undetected by the reading.

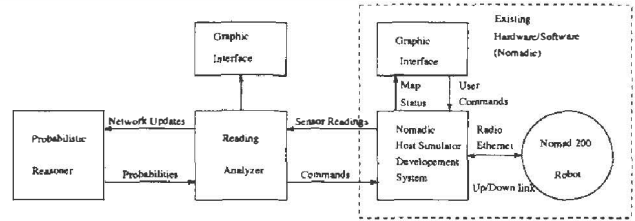

Figure 4: Architecture of the experimental system

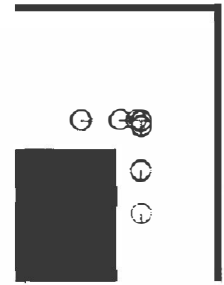

Figure 5: Scenario 1: Corridor

The probabilistic reasoner constructs the Bayes net dynamically (using IDEAL [11] in LISP), according to topological data from the readings analyzer, sets up the readings as evidence, and performs belief updating. Dynamic construction of the network is by performing actions for each new input item, as follows:

1. Add-reading(name, distance): create new reading node, con node and pro node. Connect them and set conditional probabilities.

2. Add-region(name, area): create a new region node, and set up its prior probability according to its size.

3. Add-cause(region, reading, polarity): add region node as a parent of pro or con node (according to polarity) of reading. If an or-node (pro or con node) has more than 2 parents, add an intermediate or-node.

The intermediate or-nodes in add-cause are a pure implementation issue: our version of IDEAL represents all nodes as distribution arrays, which may become too large if we allowed arbitrary in-degree. Additionally, some of the algorithms we use for belief updating work better with a small in-degree.

## 4  SIMULATION RESULTS

Results presented are for a simulated environment. The maximal reflection angle (critical angle) was set at $60^0$ for the simulation. We compare the results of our mapping to results that might be obtained by assuming grid-point independence (the Elfes model), in several spatial configurations defined with the simulator's polygon mode editing. The complexity of the network generated for a large number of sonar readings forced us, in the current implementation, to select a particular mix of readings that does not generate too many regions, but with some dropouts, in order to show that our model can discount them. Hence, our experimental setting is somewhat contrived. The configurations selected for the paper are 1) a turn in a corridor (figure 5), and 2) an irregular polygon (figure 6) where the simulated robot's path shown as a sequence of circles.

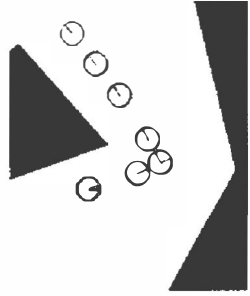

Figure 6: Scenario 2: Irregular Polygon

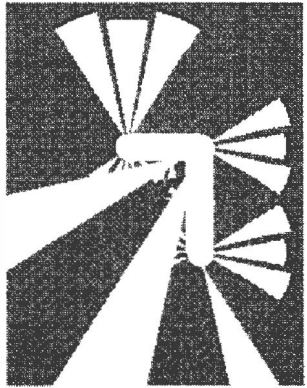

Figure 7: Scenario 1: Results (Independence)

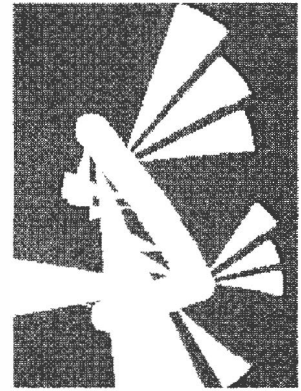

Figure 8: Scenario 2: Results (Independence)

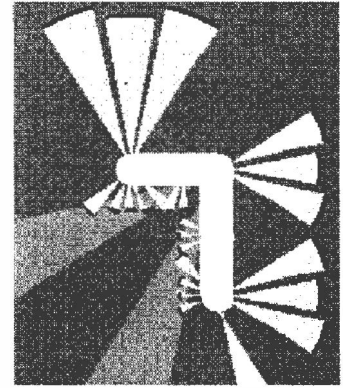

Figure 9: Scenario 1: Results (Bayes Network)

Results are presented as an array of gray-level pixels, with white representing a probability of 0 (for an obstacle), and black a probability of 1, on a linear scale. The relative grid point scale is 1 inch per grid point. Sonar arc thickness (error) width was 3 pixels.

In displaying results for independent updates (Figures 7,8) we show both the sonar readings, and the occupancy probability computed with the grid-point independence assumption, using equation 1, with initial probability being $p_{min}=0.3$. If the grid point is in the detection arc of the reading, we used $p(r \mid O) = 0.9$, and $p(r \mid \neg O) = 0.3$. If the grid point is in the "free" sector of a reading, we used $p(r \mid O) = 0.1$ and $p(r \mid \neg O) = 0.9$. Parameters for our model were $P_r = 0.9$ (the probability of true positive), prior probability of a dropout, $P_s = 0.1$, $P_0 = 0.05$, and $p_{max}$ was set at 0.8. Grid points traversed by the robot body have $P(free) = 1$.

For comparison, region occupancy probabilities in our model (which stand for probability that there is *some*

obstacle *anywhere within the region*) must be translated into a probability that a single grid point be occupied. For any region $A$, let $P(\neg free(A))$ be its prior probability of occupancy (from equation 4), and $P_\mathcal{E}(\neg free(A))$ be its posterior probability of occupancy given the evidence (the readings). Probability of occupancy for a grid point $g$ in region $A$ is given by:

$$P(occupied(g)) = p_{min} \frac{P_\mathcal{E}(\neg free(A))}{P(\neg free(A))}$$

The above equation is only for comparative display of the results. We do not intend to argue that it represents the actual probability that the (region of space covered by) a grid point is occupied given the evidence.

For scenario 1 (Figure 5), we have 26 readings, resulting in 73 regions excluding robot body, with 69 positive area regions. The network had 104 support arcs. Results are depicted in figures 9, and 10 (thresholds on figure 9, set at $P(occupied(g)) = 0.25$ and 0.35).

For scenario 2 (Figure 6), we have 20 readings. This resulted in 83 regions, excluding regions covered by the robot body, and 65 regions after removing regions including no grid points. The generated Bayes network had 139 support arcs. Results are depicted in figures 11, and 12.

With the independence assumption, several good read-

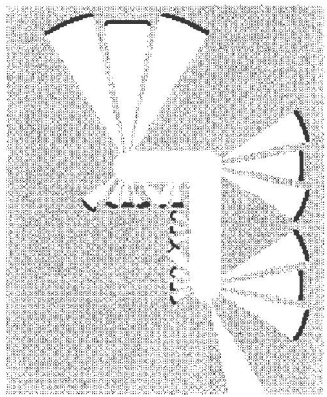

Figure 10: Scenario 1: Results with Threshold

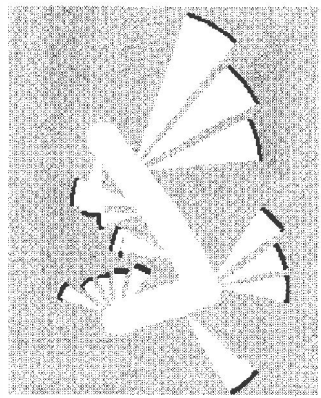

Figure 12: Scenario 2: Results with Threshold

## 5 DISCUSSION

It is possible to add prior information, or information from other sensors, into this system. One way to do this is by using different priors for regions (and other parameters than just size to decide priors). Another way is to add a prior dependency model on the regions. In order to incorporate prior knowledge of *objects* into the map (e.g. a wall known to be at a certain position, with some given probability), one could further partition regions to accommodate such known objects.

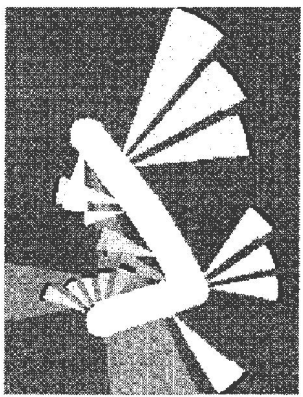

Figure 11: Scenario 2: Results (Bayes Network)

One weakness of our system is in coping with registration or location errors: if the location of a *known* object is incorrect, incorporating that is somewhat difficult. That is because the system cannot tell where region partitions should be added. Although one possible solution would be to add a fine partition, catering for all possible placements of the object, this would clearly by inefficient. A similar problem is encountered if the location of the sensors (w.r.t. earlier placement of the sensors) is known imprecisely. In both cases, the problem is not severe if the error is small. Using localization schemes available in the literature one can solve the latter problem. Handling the location errors for other objects is still an open problem.

ings are essentially disregarded (occupancy within the detection arc less than the global prior, $p_{min}$) due to one or more dropouts. The system would thus tend to ignore some nearby obstacles. Our model overcomes dropouts in both scenarios: probability of region occupancy higher than prior, despite the dropouts. The latter are much discounted, as shown by applying of thresholds to the results. This property occurs due to the fact that detection arcs of different readings (even disjoint ones), occuring wholly or partially within a "free" sector (of a dropout), tend to support each other in the model, and to *decrease* the probability that the "free" sector of the dropout is, indeed, free.

Evaluation time for the above Bayes networks was about 1 CPU-minute on a Sparc ELC, with the Jensen (junction-tree) algorithm in IDEAL. However, with more readings, or readings with a larger tendency to intersect, computation time grows rapidly. A configuration (not shown) with 25 readings and 115 regions, and took 4 hours CPU time. Larger configurations could not be evaluated at all. This problem is partially addressed below.

Currently, the system handles only the "gross" uncertainty resulting from sonar readings, caused by the beam-width and secondary reflections. It would be desirable to incorporate handling of the "small" errors resulting from inexact readings, which are handled by various sensor fusion schemes, such as Kalman filtering. In principle, this may be done *after* the "gross" sources of error have been eliminated.

Another obvious problem is that the Bayes network quickly becomes too large for evaluation by general, exact methods. We are considering several directions for overcoming the problem, as follows:

1. Using a scheme similar to that used in [1] for the somewhat similar BNO2 networks.

2. Make decisions about regions $A$ that have extreme probabilities $P(\text{free}(A))$, after a few read-

ings, thus partitioning, as well as decreasing the size of the network.

3. Using order of magnitude probabilities as an approximation.

4. Using sampling algorithms (e.g. [6, 10]). Initial experiments in [10] show some measure of success with this method.

Numerous other papers have time-dependent or dynamic Bayes network for handling problems of uncertainty in sensing and results of actions. They all, however, deal with different kinds of sensors and environments. For example, [8] has a discrete spatial representation in terms of "rooms" where light beam sensors are placed at the doors, to detect crossing objects. Discrete locations of moving objects are then estimated by dynamically constructing a Bayes network and evaluating it. Sensors may either fail to detect objects, or detect "ghosts". The spatial configuration in that paper is essentially reduced to a topology, whereas in our paper, the spatial geometry is important. Note that in our paper, the grid is used just as a convenient computational aid, and in fact we could have used a continuous geometrical representation of regions. The latter is not possible in [8].

Other examples are in traffic scene analysis [7], which uses Bayes networks to track discrete objects (cars), and in the BATmobile autonomous vehicle simulation [4] which uses Bayes networks to decide the position of obstacles (e.g. other cars). In both papers, the inherent assumption is that we can separate out the individual objects, which is indeed possible due to e.g. relative motion. That is not possible in our domain, and hence we need to use a different sensor model.

# 6  SUMMARY

This paper presents a probabilistic model for "gross" sonar sensor fusion, that relaxes several independence assumptions made in related work. Such assumptions may be over-optimistic for various environments. The dependencies we introduced between readings are represented by the topology of a (dynamically constructed) Bayes network, which also represents the conditional distributions between elements of the model (readings and spatial regions). Initial experiments indicate that the model can overcome spurious sonar reflections, by accumulating evidence from a number of readings, and thus greatly discount the impact of the bogus readings. This evidence accumulation is done automatically by the probability updating mechanism for the probability model (as implemented by the Bayes network). We believe that the method is reasonably easy to extend to incorporating prior information, both certain and uncertain, as well as information from other types of sensors.

**Acknowledgments**

This research is partially supported by the Paul Ivanier Center for Robotics and Production Management, Ben-Gurion University.